\documentclass[10pt,twocolumn,letterpaper]{article}

\usepackage{cvpr}              

\usepackage{graphicx}
\usepackage{amsmath}
\usepackage{amssymb}
\usepackage{booktabs}
\usepackage{algorithm}
\usepackage{algpseudocode} 
\usepackage{amsmath}
\usepackage{mathtools}
\usepackage{tabularx}
\usepackage{subcaption}
\usepackage{caption}
\usepackage{makecell}
\usepackage{marvosym}
\usepackage[accsupp]{axessibility}  
\usepackage[table]{xcolor}
\usepackage{url}

\newcommand{\defeq}{\vcentcolon=}

\usepackage[pagebackref,breaklinks,colorlinks]{hyperref}

\usepackage[capitalize]{cleveref}
\crefname{section}{Sec.}{Secs.}
\Crefname{section}{Section}{Sections}
\Crefname{table}{Table}{Tables}
\crefname{table}{Tab.}{Tabs.}

\def\cvprPaperID{4120} 
\def\confName{CVPR}
\def\confYear{2023}

\definecolor{black}{RGB}{0,0,0}
\definecolor{darkgray}{RGB}{100,100,100}
\definecolor{gray}{RGB}{150,150,150}

\begin{document}

\title{Shepherding Slots to Objects:\\Towards Stable and Robust Object-Centric Learning
}

\author{%
Jinwoo Kim$^1$$^*$ \quad Janghyuk Choi$^2$$^*$ \quad Ho-Jin Choi$^2$ \quad Seon Joo Kim$^1$   \\
$^1$Yonsei University \quad $^2$KAIST \\
\texttt{\small{\{jinwoo-kim, seonjookim\}@yonsei.ac.kr}}\\
\texttt{\small{{\{janghyuk.choi, hojinc\}@kaist.ac.kr}}}
}

\maketitle
\def\thefootnote{*}\footnotetext{Equal contribution, a coin is flipped.}\def\thefootnote{\arabic{footnote}}

\begin{abstract}
   Object-centric learning (OCL) aspires general and compositional understanding of scenes by representing a scene as a collection of object-centric representations.
   OCL has also been extended to multi-view image and video datasets to apply various data-driven inductive biases by utilizing geometric or temporal information in the multi-image data.
   Single-view images carry less information about how to disentangle a given scene than videos or multi-view images do.
   Hence, owing to the difficulty of applying inductive biases, OCL for single-view images remains challenging, resulting in inconsistent learning of object-centric representation.
   To this end, we introduce a novel OCL framework for single-view images, SLot Attention via SHepherding (SLASH), which consists of two simple-yet-effective modules on top of Slot Attention.
   The new modules, Attention Refining Kernel (ARK) and Intermediate Point Predictor and Encoder (IPPE), respectively, prevent slots from being distracted by the background noise and indicate locations for slots to focus on to facilitate learning of object-centric representation.
   We also propose a weak semi-supervision approach for OCL, whilst our proposed framework can be used without any assistant annotation during the inference.
   Experiments show that our proposed method enables consistent learning of object-centric representation and achieves strong performance across four datasets.
   Code is available at \url{https://github.com/object-understanding/SLASH}.
\end{abstract}

\section{Introduction}
\label{sec:intro}

\textit{Object-centric learning} (OCL) decomposes an image into a set of vectors corresponding to each distinct object to acquire object-wise representations \cite{greff2016tagger}.
Learning object-centric representation enables machines to perceive the visual world in a manner similar to humans. We recognize the world as a composition of \textit{objects} \cite{kahneman1992reviewing} and extend the object-related knowledge to various environments \cite{spelke2007core}.
Therefore, OCL enables a compositional understanding of an image and generalization for downstream tasks, such as visual reasoning \cite{mao2019neuro} and object localization \cite{cho2015unsupervised}.

\begin{figure}
  \centering
  \includegraphics[width=\linewidth]{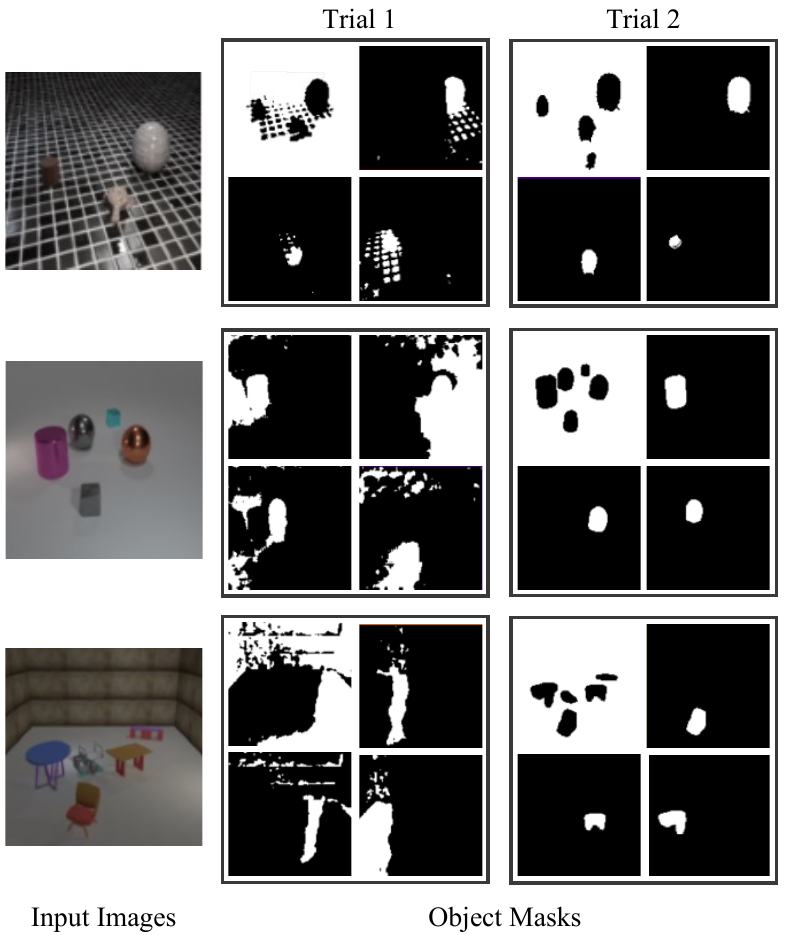}
  \caption{
  Results of training Slot Attention~\cite{locatello2020object} with different seeds, which show inconsistent learning results.
  In the first trial, object-centric representations fail to grasp each distinct object due to the background noise.
  In the second, the model succeeds in distinguishing each different object from the background.
  }
  \label{fig:teaser}
\end{figure}

Mainstream OCL has adopted an autoencoding-based compositional generative model \cite{greff2019multi, locatello2020object, engelcke2019genesis}.
Slot Attention \cite{locatello2020object} is the most prominent technique for OCL, which uses \textit{slots} as the intermediate representation bottlenecks.
In the Slot Attention, randomly initialized slots compete with each other to occupy their attention regions in terms of pixels.
Eventually, each slot attains object-centric representation by aggregating visual features according to the attention map between the slot and pixels.

Recently, OCL has been extended to multi-view images \cite{chen2021roots, sajjadi2022object} and videos \cite{singh2022simple, kipf2021conditional, elsayed2022savi++}.
Multi-view image \cite{sajjadi2022scene} or video \cite{girdhar2020cater, sun2020scalability, greff2021kubric} datasets allow models to learn spatial geometry or temporal dynamics of objects through supplementary objective tasks such as novel view synthesis \cite{sajjadi2022object} and optical flow inference \cite{kipf2021conditional}.
Consequently, these datasets provide additional information that enables the adoption of data-driven inductive biases, facilitating the learning of better object-centric representations. 

In contrast, it is challenging to obtain data-driven inductive biases, such as geometric or temporal information, for single-view images. 
To address this problem, novel architectures, such as auto-regressive generative models \cite{burgess2019monet, greff2019multi, engelcke2019genesis, engelcke2021genesis} and Transformer \cite{vaswani2017attention} for encoders \cite{seitzer2022bridging} and decoders \cite{singh2021illiterate}, have been proposed. 
However, owing to the absence of additional inductive biases, OCL for complex single-view images suffers from unstable training results.

This stability issue implies inconsistent learning of object-centric representation, that is, not all trials of training a model with the same architecture consistently succeed in distinguishing objects from the background (\cref{fig:teaser}).
The attention-leaking problem, or bleeding issue, can mislead a model to yield object-centric representations based on distorted attention maps.
The bleeding issue is fatal for OCL because it is difficult to predict the behavior of a model, that is, whether a slot will seize a distinct object or an object entangled with a background.

To solve this bleeding issue, we propose a novel OCL framework, SLASH (\textbf{SL}ot \textbf{A}ttention via \textbf{SH}epherding).
SLASH resolves the bleeding by guiding the randomly initialized slots to successfully grasp objects 1) without being distracted by the background and 2) by keeping informed of the destination.
These are accomplished by adding two simple-yet-effective modules, Attention Refining Kernel (ARK) and Intermediate Point Predictor and Encoder (IPPE), to the Slot Attention framework.

ARK is a single-channel single-layer convolutional kernel, designed to prevent slots from focusing on a noisy background.
We adopt the Weights-Normalized Convolutional (WNConv) kernel, a learnable low-pass filter, as the kernel for ARK. 
This simple kernel refines the attention map between slots and pixels by reducing noise and solidifying object-like patterns.

IPPE serves as an indicator to nudge a slot to focus on the proper location.
Thus, the slots can consistently update their representations without being confused by the background.
IPPE consists of two submodules with simple MLPs.
The first submodule predicts the position of an object in two-dimensional coordinates, and the second encodes the predicted coordinates into a high-dimensional vector.

Since IPPE needs to be trained to provide locational cues to slots, it is necessary to introduce positional labels.
However, using fully annotated ground-truths is costly, particularly for densely-annotated labels such as object masks.
Hence, we adopt a weak semi-supervision approach in which only a small subset of the dataset includes weak annotations, such as the centers of bounding boxes. 
We show that IPPE can be successfully trained with weakly semi-supervised learning and can be deployed under circumstances where no assistant ground-truth exists.

For a comprehensive study, we validate our method on numerous datasets, including CLEVR, CLEVRTEX, PTR, and MOVi.
Moreover, we conduct 10 trials of training for each method, including the baselines and ours, to thoroughly evaluate the results.
We estimate the performance of the models using three metrics: mean Intersection over Union (mIoU), Adjusted Rand Index (ARI), and foreground-ARI (fg-ARI)
In particular, mIoU and ARI investigate whether the bleeding issue occurs by considering the background separation.
A model is defined as being stable over the metrics when deviations are lower, and as being robust when averages are higher across all datasets.
Experimental results demonstrate that our method achieves stable and robust OCL that prevents the bleeding issue.

Our main contributions are as follows:
\begin{itemize}
    \item We observe OCL for single-view images suffers from the stability issue with inconsistent training results. To resolve this issue, we propose a novel framework, SLASH (\textbf{SL}ot \textbf{A}ttention via \textbf{SH}epherding) consisting of two simple-yet-strong modules: ARK and IPPE.
    \item ARK is a learnable low-pass filter designed to prevent the bleeding issue where the attention of a slot leaks into a background.
    \item IPPE is introduced to inform slots of the regions to be focused. By leveraging weak semi-supervision, IPPE can inject positional information into a slot.
    \item We empirically prove SLASH achieves stable and robust OCL against four distinctive datasets. SLASH shows the best stability while outperforming the previous methods for all datasets over multiple metrics.   
\end{itemize}


\section{Related Works}
\label{sec:related_works}
\subsection{Object-Centric Representation Learning}
\label{subsec:related_works_ocl}
A line of works for OCL adopts the scene reconstruction, where a model learns to decompose an image into several components without using any human-annotated ground-truths \cite{stelzner2019faster, greff2016tagger, greff2017neural, eslami2016attend, crawford2019spatially, jiang2019scalor, lin2020space, prabhudesai2022generating}. 
MONet \cite{burgess2019monet} and IODINE \cite{greff2019multi} proposed unsupervised auto-regressive approaches to sequentially disentangle object-centric representations from a scene.
GENESIS \cite{engelcke2019genesis, engelcke2021genesis} improved object-centric learning by enabling interactions between slots while using an auto-regressive approach.
Slot Attention \cite{locatello2020object} introduced an attention-based mechanism between slots and pixels, where slots parallelly and iteratively compete with each other to occupy their own territory in the pixel space.
Slot Attention improved training speed and memory efficiency by enabling the parallel update of slots.
Recently, SLATE \cite{singh2021illiterate} and DINOSAUR \cite{seitzer2022bridging} adopted Transformer \cite{vaswani2017attention} as an encoder and decoder for Slot Attention, respectively, to learn object-centric representations over real-world images.

Several studies have adapted \textit{novel view synthesis} (NVS) to OCL \cite{chen2021roots, stelzner2021decomposing, henderson20neurips, sajjadi2022object}.
ROOTS \cite{chen2021roots} proposed an approach to infer 3D disentangled object representation using 3D-to-2D perspective projection \cite{hartley2003multiple} with multi-view images.
Other studies \cite{sajjadi2022object, stelzner2021decomposing} directly applied Slot Attention for multi-view images and demonstrated that using multi-view images with NVS significantly improves OCL performance.

OCL for videos has been actively studied \cite{van2018relational, jiang2019scalor, kabra2021simone, kipf2021conditional, elsayed2022savi++, singh2022simple}. SAVi \cite{kipf2021conditional}, SAVi++\cite{elsayed2022savi++} and STEVE \cite{singh2022simple} extended Slot Attention to videos, in which a model iteratively infers object-centric representations across a sequence of images.
With a sequence of images, models can learn to distinguish objects from backgrounds by referring to the temporal consistency and dynamics of the objects.
In this study, we focus on a more challenging case, OCL for single images, where less information about an object and its background is provided than in OCL for multi-view images and videos. 

\begin{figure*}
  \centering
  \includegraphics[width=0.95\linewidth]{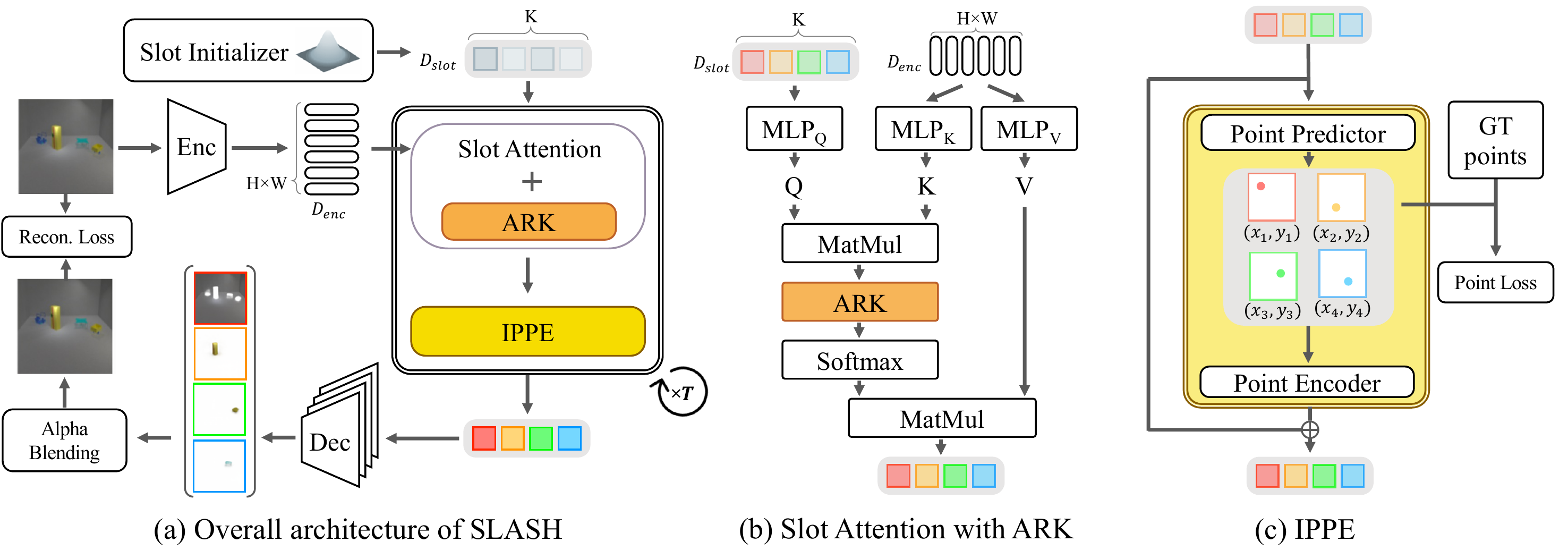}
   \caption{
    (a) Overall architecture of the proposed framework. Upon Slot Attention (modules without color fillings), we add Attention Refining Kernel (ARK, filled with orange color) and Intermediate Point Predictor and Encoder (IPPE, filled with yellow color).
    (b) Within the Slot Attention module, we insert ARK before the softmax function.
    (c) IPPE predicts 2D point coordinates and encode the coordinates into vectors of dimension $D_{slots}$. The point-encoded vectors are then added to the slots so that the slots can incorporate position information.
   }
   \label{fig:model}
\end{figure*}

\subsection{Weakly Supervised OCL}

In weakly supervised learning, training is conducted with human-annotated labels that provide insufficient or indirect information but are pertinent to obtaining the target outputs. 
In OCL, GFS-Net \cite{prabhudesai2022generating} viewed the learning of object representations as a combination of \textit{what and where} problem.
GFS-Net is first trained with images containing only a single object and then fine-tuned with images containing multiple objects to resolve the what and where problem.
PriSMONet \cite{elich2022weakly} used shape priors in weakly supervised learning for multi-object 3D scene decomposition over RGB-D images.
Furthermore, in OCL for videos, SAVi \cite{kipf2021conditional} and SAVi++ \cite{elsayed2022savi++} used position information, such as the center of a mass, bounding box, or object mask, for each object in the first frame of the given video to deal with the video-level OCL.
In this work, motivated by \cite{kipf2021conditional, elsayed2022savi++}, we utilize the point information of objects in a scene to iron out the subjectiveness problem in the image-level OCL.

\subsection{Semi-Supervised Learning}

In deep learning studies, any type of human-annotated labels, even with coarse or sparse information, can help enhance the performance of models. 
However, it is unfeasible to place ground-truths everywhere during the training and testing phases.
Several semi-supervision studies have investigated how to leverage the lack of labels to solve image classification \cite{laine2016temporal, haeusser2017learning, berthelot2019mixmatch, oh2021distribution}, object detection \cite{chen2021points, xu2021end, sohn2020simple, jeong2019consistency}, and semantic/instance segmentation \cite{papandreou2015weakly, khoreva2017simple, ahn2018learning, lai2021semi} problems.

In this study, we adopt a novel approach for OCL, where models can only use weak supervision labels for a fraction of a given dataset.
The most comparable study for different tasks is Point DETR \cite{chen2021points}.
Point DETR focused on weakly semi-supervised object detection using a dataset consisting of a few fully-annotated images with bounding boxes and object category labels, and a rich amount of weakly-annotated images with center points and object category labels.
However, instead of fully-annotated images with semantic labels, our method only uses a few amounts of weakly-annotated images with point-level labels and a significant amount of non-annotated images.

\section{Method}

\subsection{Preliminary: Slot Attention}
In Slot Attention \cite{locatello2020object}, the object-centric representation is implemented with the concept of \texttt{slots} $\in \mathbb{R}^{K \times D_{slot}}$, which is a set of $K$ vectors of dimension $D_{slot}$.
The slots are initialized by a Gaussian distribution with a learnable mean $\mu$ and sigma $\sigma$, and updated over $T$ iterations by the Slot Attention module.
The slots are then decoded into the target reconstruction image.

We first describe the overall procedure of how Slot Attention is trained for the completeness of this study.
Given an image, a convolutional neural network (CNN) encoder produces a visual feature map of dimension $HW \times D_{enc}$, where $H$ and $W$ are the height and width of an input image.
The Slot Attention module takes \texttt{slots} and the visual feature map, called \texttt{inputs}, then projects them to dimension $D$ by a linear transformation $k$ for slots and $q$, $v$ for \texttt{inputs} $\in \mathbb{R}^{HW \times D_{enc}}$.
Dot-product attention is applied to generate an attention map, \texttt{attn}, with query-wise normalized coefficients where slots compete with each other to occupy the more relevant pixels of the visual feature map (\cref{eq1} \cite{locatello2020object}).

\begin{equation}\label{eq1}
\begin{gathered}
\texttt{attn}_{i,j} \defeq \frac{\exp(M_{i,j})}{\Sigma_{l} \exp(M_{i,l})}, \quad where\\ 
M \defeq \frac{1}{\sqrt{D}}k(\texttt{inputs}) \cdot q(\texttt{slots})^{T} \in \mathbb{R}^{HW \times K}.
\end{gathered}
\end{equation}

The projected visual feature map weighted by the attention map coefficients (\cref{eq2} \cite{locatello2020object}) is aggregated to produce the updated slots, \texttt{updates}. 
As the Slot Attention module runs iteratively, \texttt{slots} can gradually update their representation.
Each updated slot is then decoded into an RGB-A image using a spatial broadcast decoder \cite{watters2019spatial}, where the weights are shared across slots.
The decoded images are blended into a single image using alpha masks to yield the final reconstructed image.
The mean squared error (MSE) between the original input image and the predicted reconstruction image is chosen for the objective function so that the overall training follows unsupervised learning.

\begin{equation}\label{eq2}
\begin{gathered}
\texttt{updates} \defeq W^{T} \cdot v(\texttt{inputs}) \in \mathbb{R}^{K \times D}, \\
where \quad W_{i,j} \defeq \frac{\texttt{attn}_{i,j}}{\Sigma_{l=1}^{N}\texttt{attn}_{l,j}}.
\end{gathered}
\end{equation}

\subsection{SLot Attention via SHepherding (SLASH)}
In this work, our goal is to achieve stable and robust OCL in single-view images by preventing the bleeding issue incurred when slots are distracted by background noise.
To achieve this goal, the model needs to provide guidance to the slots about where to focus or not. 
To this end, we propose a novel OCL framework, SLASH (\textbf{SL}ot \textbf{A}ttention via \textbf{SH}epherding), which steers slots to correctly seize objects using two newly introduced modules: Attention Refining Kernel (ARK) and Intermediate Point Predictor and Encoder (IPPE).
ARK guards and stabilizes slots against background noise by reducing the noise and solidifying object-like patterns in the attention map between the slots and pixels.
IPPE guides slots towards the area where an object is likely to exist by providing positional indications to the slots. 
Using these two simple-yet-effective modules that shepherd slots to the desired region, SLASH accomplishes stable and robust OCL.
The overall architecture of SLASH is shown in \cref{fig:model}.

\subsubsection{Attention Refining Kernel}
\label{subsubsec:ark}
\begin{figure}
  \centering
  \includegraphics[width=\linewidth]{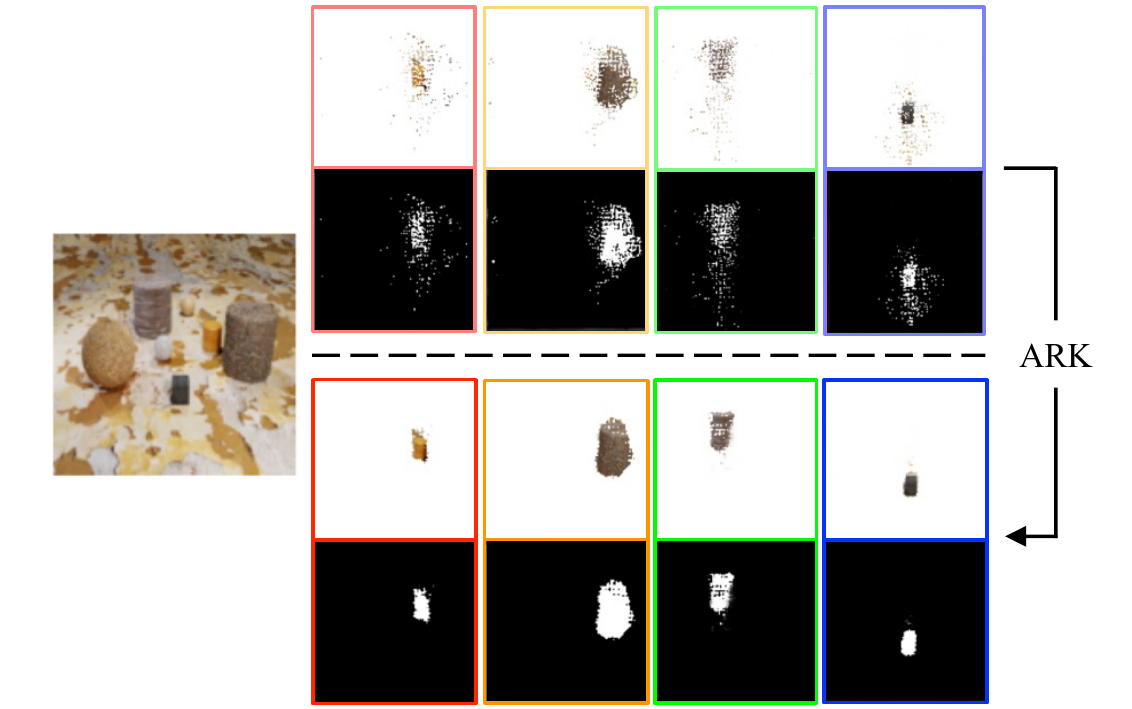}
  \caption{
  Visualization of the results by Attention Refining Kernel (ARK).
  Each colored box shows an attention map between a slot and pixels.
  The upper part of a rectangular box is a visualization of the attention map on the input image.
  The lower part represents the attention map in grayscale. 
  The top and bottom row, split by the dotted line, corresponds to the attention map before and after applying ARK, respectively.
  One can observe that ARK refines the scattered attention around objects so that slots can escape from the background noise. 
  }
  \label{fig:ark}
\end{figure}

Attention Refining Kernel (ARK) is designed to prevent slots from being distracted by background noise by refining the attention map between slots and visual features.
As depicted in the upper part of \cref{fig:ark}, we can observe that Slot Attention \cite{locatello2020object} generates attention maps with salt-and-pepper-like noise around the objects.
Noisy attention maps are likely to provoke unstable learning of object-centric representations.
We address this issue by introducing an inductive bias for \textit{local density} of objects.
The bias for local density assumes that the density of the attention values should be higher near an object and lower outside the object.
Thus, the inductive bias is materialized using the Weights-Normalized Convolutional (WNConv) kernel which aims to refine an attention map by reducing noise and solidifying object-like patterns around objects.
WNConv kernel is a single-channel single-layer convolutional network trained under the constraints that the sum of the kernel weights equals $1$ while maintaining every weight greater than or equal to $0$.
With these constraints, the WNConv kernel serves as a low-pass filter, smoothing the attention map as shown in the lower part of \cref{fig:ark}. 
As depicted in \cref{fig:model} (b), ARK is applied to the logit values of the attention map.

\subsubsection{Intermediate Point Predictor and Encoder}
\label{subsubsec:ippe}
Intermediate Point Predictor and Encoder (IPPE) expedites learning ``where objects exist''.
In order for IPPE to understand the location of objects, it is necessary to introduce external supervision related to the object positions.
To train our model practically, we utilize a weak semi-supervision approach.
We use the low-cost information, center points of bounding boxes, as the weak supervision among the possible positional cues.
Furthermore, instead of using a fully annotated dataset, we assume that only a fraction (10\%) of the dataset and not all objects in a given image (75\%) have labels.
The following describes how IPPE leverages weak semi-supervision.

IPPE consists of two modules, a point predictor and a point encoder, as shown in \cref{fig:model} (c).
The point predictor is a 3-layer MLP that predicts 2D point coordinates of objects from slots.
The point encoder, also a 3-layer MLP, encodes the point coordinates into $D_{slot}$ dimensional vectors, which are added to the original slots.
The updated slots can now contain information about the location of objects and become less likely to wander around the background.

\begin{figure}
  \centering
  \includegraphics[width=\linewidth]{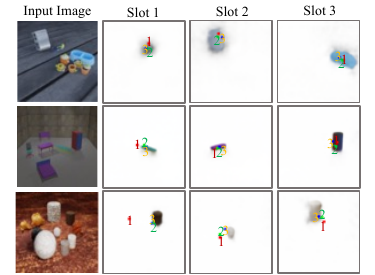}
  \caption{Visualization of point predictions by Intermediate Point Predictor and Encoder (IPPE).
  The leftmost column shows the input images, and the right three columns show the prediction results by IPPE for each slot.
  Each number stands for the order of iteration $T$.
  Best viewed in color.
  }
  \label{fig:ippe}
\end{figure}

The point predictor is trained by weak semi-supervision with an auxiliary point loss which is MSE between the predicted and ground-truth coordinates.
Hungarian algorithm \cite{kuhn1955hungarian} is used to match the predictions and ground-truths.
\cref{fig:ippe} shows the results of the point predictor, where the predictions get closer to objects through slot updates.

The point encoder is trained using the image reconstruction loss in a self-supervised manner \cite{locatello2020object}.
As the reconstruction loss is shared with the Slot Attention module, the point encoder generates position-encoded vectors that are well-aligned with the Slot Attention module.
It is worth noting that the point encoder can take either ground-truths as inputs, if available, or the predicted coordinates from the point predictor, otherwise.

Our method differs from the previous weakly supervised OCL method for videos~\cite{kipf2021conditional} in that we use weak semi-annotations as the ground-truth labels for the module (Point Predictor) as well as the input for the module (Point Encoder).
Conversely, SAVi exploits weak supervision to initialize slots using an MLP with the ground-truth position information as its input.
That weakly supervised slot initialization shows outstanding performance for video OCL; however, it is limited in the sense that the model requires labels for all samples, even during inference time.
This limitation arises from the lack of preparation for the cases where the ground-truths are not provided or are partially provided in both the training and inference phases.
By virtue of the design of IPPE, our method can be trained in a weakly semi-supervised manner and can be deployed under circumstances where no ground-truth exists.
In the following section, we validate the proposed method against a SAVi-like OCL method for images.

\section{Experiments}

\subsection{Experimental Setup}
\label{subsec:exp_setup}

\noindent
\textbf{Task \& Metrics} \quad
To validate the effectiveness of our method, we conduct experiments on the object discovery task following the previous OCL works \cite{locatello2020object, kipf2021conditional, engelcke2019genesis, engelcke2021genesis, prabhudesai2022generating, elsayed2022savi++}.
In the object discovery task, a model is required to cluster pixels into object segments.
Though the task seems similar to the instance segmentation, the object discovery differs from the image segmentation in that it does not require semantic classes or captions for each segmentation.

\begin{figure}
  \centering
  \includegraphics[width=\linewidth]{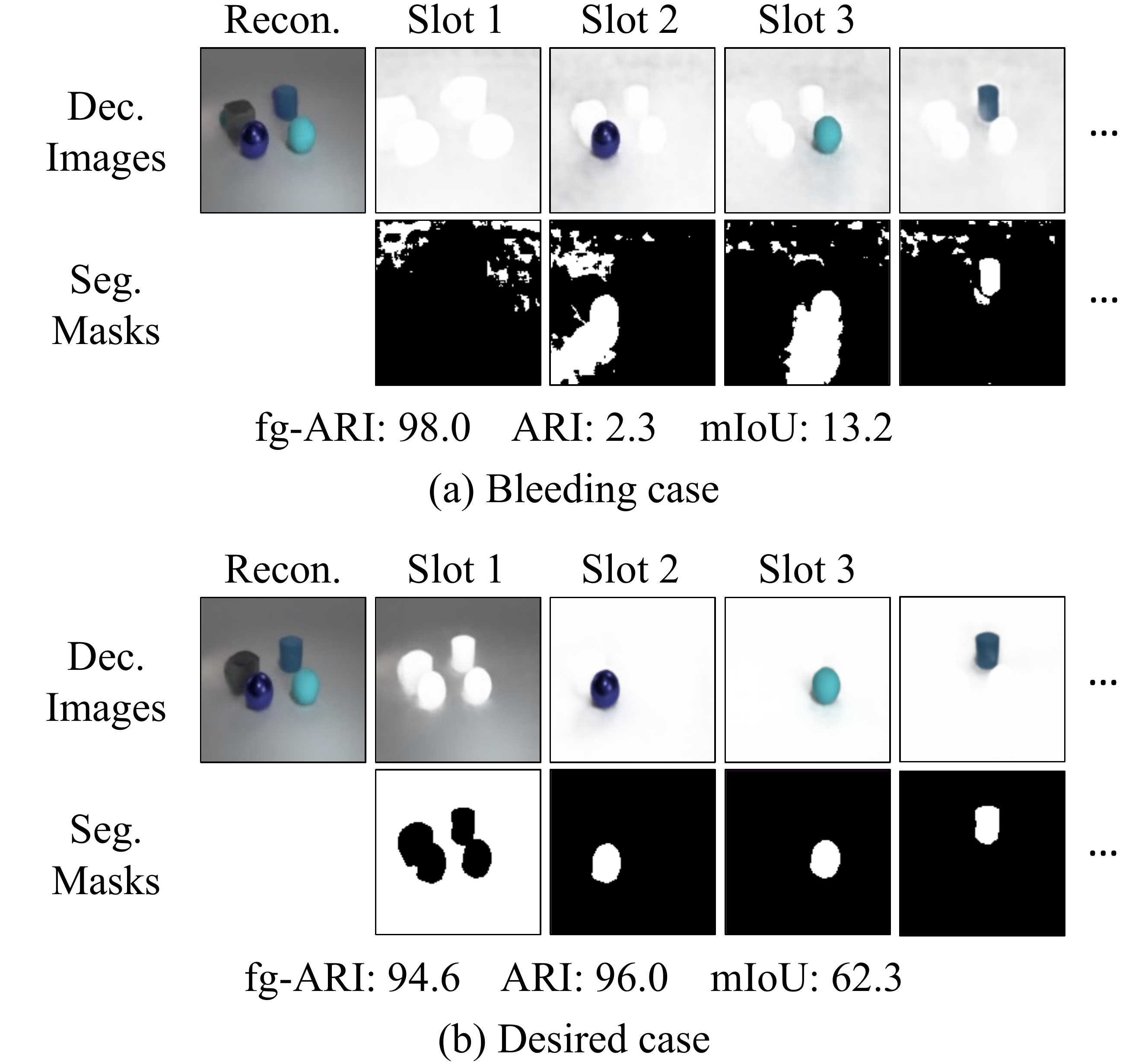}
  \vspace{-1.5em}
  \caption{
  Demonstration of the bleeding issue.
  The top row of each case shows decoded images by a model \cite{locatello2020object} and the bottom row shows segmentation masks corresponding to the decoded images.
  At the bottom of each case, evaluation results by three different metrics are reported in $\%$. 
  One can observe that fg-ARI cannot represent the bleeding case in contrast to ARI and mIoU.
  }
  \label{fig:bleeding}
\end{figure}

To evaluate the models, we use mean Intersection over Union (mIoU) and Adjusted Rand Index (ARI) \cite{rand1971objective}. 
Similar to \cite{engelcke2019genesis, monnier2021unsupervised}, we avoid focusing on foreground-ARI (fg-ARI) where the annotation for the background is excluded from the evaluation.
\cref{fig:bleeding} demonstrates that fg-ARI cannot describe the stability issue, such as the bleeding.
On the other hand, the stability issue can be demonstrated using mIoU and ARI since the annotation of backgrounds is concerned with those metrics.

\noindent
\textbf{Baselines} \quad
We compare SLASH with Slot Attention (SA) \cite{locatello2020object}, GENESIS-v2 (GenV2) \cite{engelcke2021genesis}, and weakly supervised Slot Attention (WS-SA).
GenV2 is a recent study on OCL in single-view images, derived from GENESIS \cite{engelcke2019genesis}.
The official GenV2\footnote{https://github.com/applied-ai-lab/genesis} is used in our datasets. 
Additionally, we compare SLASH with WS-SA, a simple variant of SA, equipped with a weakly supervised slot initializer.
The WS-SA initializes each slot using an MLP, which takes the point coordinates of an object as input by following SAVi \cite{kipf2021conditional}.
Unlike SAVi, we assume the datasets do not contain labels for the precise number of objects in an image.
Therefore, we initialize the surplus slots with randomly sampled values from the Gaussian distribution as opposed to SAVi which initializes surplus slots that receive no ground-truth point coordinates with $(-1, -1)$ to deactivate the slots.

\noindent
\textbf{Datasets} \quad
The experiments cover four multi-object datasets: CLEVR6 \cite{johnson2017clevr}, CLEVRTEX \cite{karazija2021clevrtex}, PTR \cite{hong2021ptr} and MOVi-C \cite{greff2021kubric}.
CLEVR was designed to assess the models’ comprehension of compositional elements, such as visual reasoning.
CLEVR6 contains 35K train and 7.5K validation samples consisting of scenes with three to six objects \cite{multiobjectdatasets19, locatello2020object}. 
CLEVRTEX is a variant of CLEVR, having complicated shapes, textures, materials, and backgrounds.
CLEVRTEX contains 50K samples, which we split into 40K train and 10K validation set.
PTR, which contains 52K train and 9K validation samples, is a visual reasoning dataset in which objects have part-whole hierarchies.
MOVi-C is a synthetic video dataset comprising realistic and textured daily objects and backgrounds.
We collected the first frames of the randomly rendered videos that have scenes with at most six objects.
Our MOVi-C dataset contains 39K train and 9K validation samples.
The supplementary material contains details of the data collection process.

\noindent
\textbf{Training} \quad
All models are trained by the MSE reconstruction loss in an autoencoding fashion. 
The training environments for WS-SA and SLASH are the same as those of SA \cite{locatello2020object}, while those of GenV2 follow the official paper \cite{engelcke2021genesis}.
The number of slots, $K$, is set to 7 for CLEVR6, PTR, and MOVi-C, and 11 for CLEVRTEX.


\begin{table}
    \begin{center}
    \resizebox{0.98\linewidth}{!}{
    \begin{tabular}{lccc}
        \Xhline{3\arrayrulewidth}
        & \multicolumn{1}{c}{mIoU} & \multicolumn{1}{c}{ARI} &  \multicolumn{1}{c}{fg-ARI} \\ 
        
        \hline\hline & \multicolumn{3}c{CLEVR6} \\ \hline
        SA \cite{locatello2020object} & 49.5 $\pm$ 20.5 & 63.0 $\pm$ 42.1 & 97.1 $\pm$ \phantom{0}1.6  \\
        SA\textdagger & 46.4 $\pm$ 24.6 & 59.2 $\pm$ 48.2 & 98.3 $\pm$ \phantom{0}0.9 \\
        WS-SA$^{*}$ & 61.8 $\pm$ \phantom{0}4.3 & 90.7 $\pm$ \phantom{0}2.0 & 93.3 $\pm$ \phantom{0}1.6 \\
        \textbf{SLASH$^{*}$} & 63.6 $\pm$ \phantom{0}4.3 & 90.3 $\pm$ \phantom{0}4.3 & 94.2 $\pm$ \phantom{0}1.3 \\
        
        \Xhline{2\arrayrulewidth} & \multicolumn{3}{c}{CLEVRTEX} \\ \hline
        SA & 22.2 $\pm$ \phantom{0}4.3 & 38.1 $\pm$ 12.5 & 52.1 $\pm$ \phantom{0}5.9 \\
        MONet \cite{burgess2019monet}\textdaggerdbl & 19.8 $\pm$ \phantom{0}1.0 & \multicolumn{1}{c}{---} & 36.7 $\pm$ \phantom{0}0.9 \\
        IODINE \cite{greff2019multi}\textdaggerdbl & 29.2 $\pm$ \phantom{0}0.8 & \multicolumn{1}{c}{---} & 59.2 $\pm$ \phantom{0}2.2 \\
        GenV2 \cite{engelcke2021genesis}\textdaggerdbl & \phantom{0}7.9 $\pm$ \phantom{0}1.5 & \multicolumn{1}{c}{---} & 31.2 $\pm$ 12.4 \\
        WS-SA$^{*}$ &  22.4 $\pm$ \phantom{0}4.5 & 36.0 $\pm$ \phantom{0}7.2 & 52.3 $\pm$ \phantom{0}7.3 \\
        \textbf{SLASH$^{*}$} & 34.7 $\pm$ \phantom{0}5.3 & 59.4 $\pm$ 11.5 & 61.9 $\pm$ \phantom{0}6.4 \\
        
        \Xhline{2\arrayrulewidth} & \multicolumn{3}{c}{PTR} \\ \hline
        SA & 17.6 $\pm$ 14.7 & 19.6 $\pm$ 29.8 & 44.5 $\pm$ 18.8 \\
        GenV2 & 28.5 $\pm$ 11.3 & 41.0 $\pm$ 25.4 & 56.8 $\pm$ 13.0 \\
        WS-SA$^{*}$ & 23.8 $\pm$ 15.5 & 21.4 $\pm$ 33.6 & 52.9 $\pm$ 11.6 \\
        \textbf{SLASH$^{*}$} & 44.1 $\pm$ \phantom{0}9.6 & 67.9 $\pm$ 22.6 & 59.0 $\pm$ \phantom{0}3.2 \\
        \Xhline{2\arrayrulewidth} & \multicolumn{3}{c}{MOVi} \\ \hline
        SA & 23.0 $\pm$ \phantom{0}9.8 & 25.9 $\pm$ 20.3 & 48.7 $\pm$ \phantom{0}7.0 \\
        GenV2 & 10.8 $\pm$ \phantom{0}1.1 & \phantom{0}3.6 $\pm$ \phantom{0}0.2 & 47.1 $\pm$ \phantom{0}5.8 \\
        WS-SA$^{*}$ & 21.6 $\pm$ 11.5 & 22.8 $\pm$ 21.3 & 46.2 $\pm$ \phantom{0}8.5 \\
        \textbf{SLASH$^{*}$} & 27.7 $\pm$ \phantom{0}5.9 & 34.6 $\pm$ 13.5 & 51.9 $\pm$ \phantom{0}4.0 \\
        \Xhline{3\arrayrulewidth}
    \end{tabular}
    }
    \end{center}
    \vspace{-1.5em}
    \caption{
    Results over the object discovery task (mean $\pm$ std for 10 trials, reported in $\%$).
    * indicates that the model is trained by weakly semi-supervised learning.
    All models performed inference with no assistant label. 
    \textdagger\ is for the results by \cite{locatello2020object} which uses a center crop.
    \textdaggerdbl\ is for the results from \cite{karazija2021clevrtex} which conducted three trials of training for each method with a center crop.
    }
    \label{tab:od}
\end{table}

\subsection{Object Discovery}
\label{subsec:experiments_od}
The quantitative results on the object discovery task are summarized in \cref{tab:od}.
The bleeding case causes significant degradation of mIoU and ARI, that is, the metrics have higher deviations and lower averages.
We argue that a model is stable when it has lower deviations and robust when it has higher averages across all datasets.
Thus it is crucial to prevent the bleeding case for stable and robust OCL.
SLASH records the highest average value of mIoU and ARI for almost all datasets except for ARI on CLEVR6 with a minimal difference.
In addition, SLASH demonstrates lower standard deviation values of mIoU and ARI across overall datasets.
To sum up, SLASH scores the highest and the most consistent performance across all datasets, achieving stable and robust OCL.
We provide abundant qualitative results in the supplementary material due to spatial constraints.



\subsection{Ablation Studies}

\subsubsection{ARK and IPPE}
To prove the effectiveness of ARK and IPPE, we conduct an ablation study on those modules by training SA \cite{locatello2020object} with or without each module. 
\cref{tab:ablation_ask_ippe} demonstrates that SLASH benefits from both ARK and IPPE.

We observe that ARK apparently stabilizes the model, resulting in low standard deviation values for overall datasets.
IPPE boosts the performance of both SA and `+ARK'.
Although the standard deviation values tend to be high due to the absence of ARK, we find that IPPE aids in learning against a complicated dataset, i.e. MOVi, where slots struggle to grasp the visual patterns of objects.
We argue that the positional information given by IPPE can aid the slots in binding appropriate objects more effectively.

\begin{table}
\begin{center}
    \resizebox{0.95\linewidth}{!}{
    \begin{tabular}{lccc}
        \Xhline{3\arrayrulewidth}
        & \multicolumn{1}{c}{mIoU} & \multicolumn{1}{c}{ARI} &  \multicolumn{1}{c}{fg-ARI} \\ 
        \hline\hline & \multicolumn{3}c{CLEVR6} \\ \hline
        SA \cite{locatello2020object} & 49.5 $\pm$ 20.5 & 63.0 $\pm$ 42.1 & 97.1 $\pm$ \phantom{0}1.6  \\
        + ARK & 64.1 $\pm$ \phantom{0}3.1 & 89.9 $\pm$ \phantom{0}2.2 & 95.3 $\pm$ \phantom{0}1.3 \\
        + IPPE & 57.8 $\pm$ 14.6 & 84.9 $\pm$ 27.9 & 95.7 $\pm$ \phantom{0}1.0 \\
        + ARK + IPPE & 63.6 $\pm$ \phantom{0}4.3 & 90.3 $\pm$ \phantom{0}4.3 & 94.2 $\pm$ \phantom{0}1.3 \\
        \Xhline{2\arrayrulewidth} & \multicolumn{3}{c}{CLEVRTEX} \\ \hline
        SA & 22.2 $\pm$ \phantom{0}4.3 & 38.1 $\pm$ 12.5 & 52.1 $\pm$ \phantom{0}5.9 \\
        + ARK & 31.4 $\pm$ \phantom{0}6.6 & 55.6 $\pm$ 13.2 & 57.8 $\pm$ \phantom{0}7.7 \\
        + IPPE & 25.1 $\pm$ \phantom{0}7.4 & 40.4 $\pm$ 15.6 & 54.9 $\pm$ \phantom{0}7.3 \\
        + ARK + IPPE & 34.7 $\pm$ \phantom{0}5.3 & 59.4 $\pm$ 11.5 & 61.9 $\pm$ \phantom{0}6.4 \\
        \Xhline{2\arrayrulewidth} & \multicolumn{3}{c}{PTR} \\ \hline
        SA & 17.6 $\pm$ 14.7 & 19.6 $\pm$ 29.8 & 44.5 $\pm$ 18.8 \\
        + ARK & 43.8 $\pm$ \phantom{0}3.0 & 62.3 $\pm$ 19.4 & 60.4 $\pm$ \phantom{0}3.2 \\
        + IPPE & 38.4 $\pm$ 12.8 & 58.4 $\pm$ 31.3 & 58.5 $\pm$ \phantom{0}3.1 \\
        + ARK + IPPE & 44.1 $\pm$ \phantom{0}9.6 & 67.9 $\pm$ 22.6 & 59.0 $\pm$ \phantom{0}3.2 \\
        \Xhline{2\arrayrulewidth} & \multicolumn{3}{c}{MOVi} \\ \hline
        SA & 23.0 $\pm$ \phantom{0}9.8 & 25.9 $\pm$ 20.3 & 48.7 $\pm$ \phantom{0}7.0 \\
        + ARK  & 26.2 $\pm$ \phantom{0}6.1 & 33.2 $\pm$ 13.7 & 51.0 $\pm$ \phantom{0}3.7 \\
        + IPPE & 27.2 $\pm$ \phantom{0}7.9 & 36.2 $\pm$ 16.8 & 50.8 $\pm$ \phantom{0}5.7 \\
        + ARK + IPPE & 27.7 $\pm$ \phantom{0}5.9 & 34.6 $\pm$ 13.5 & 51.9 $\pm$ \phantom{0}4.0 \\
        \Xhline{3\arrayrulewidth}
    \end{tabular}
    }
    \end{center}
    \vspace{-1em}
    \caption{Results of the ablation studies on the modules of SLASH
    (mean $\pm$ std for 10 trials, reported in $\%$).}
    \label{tab:ablation_ask_ippe}
\end{table}

\begin{figure*}
  \centering
  \includegraphics[width=\linewidth]{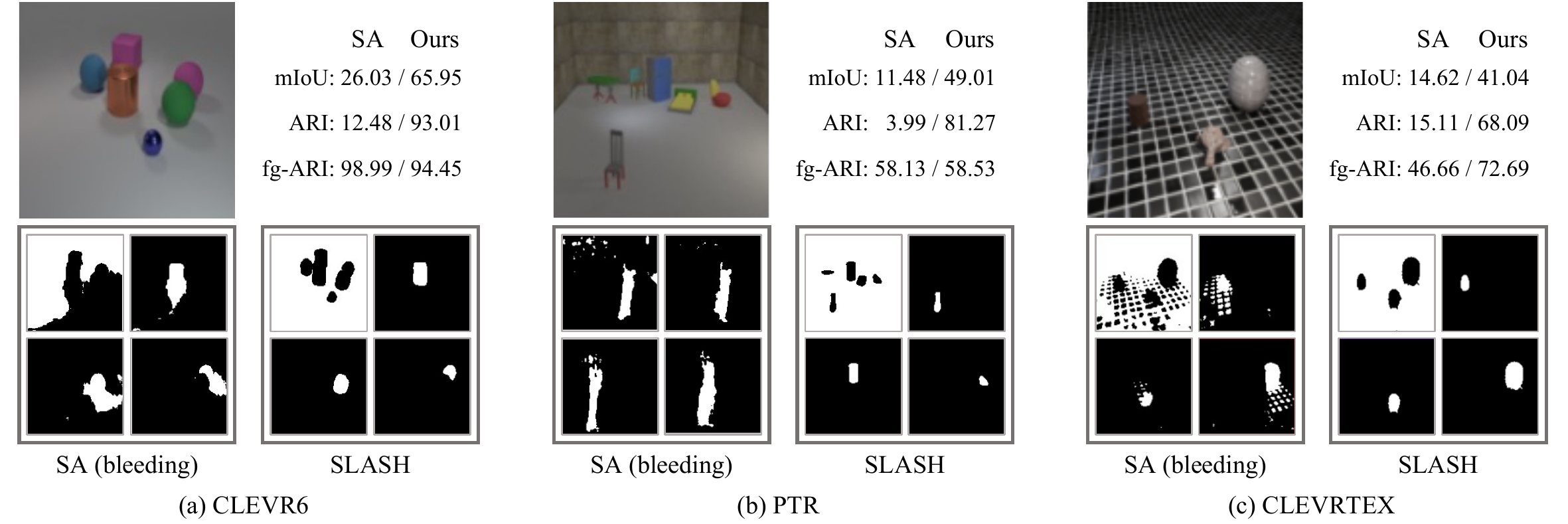}
  \vspace{-2em}
  \caption{Qualitative evaluation of SLASH compared to the baseline, Slot Attention (SA) \cite{locatello2020object}. The images in the top left of each dataset section are randomly selected inputs fed to the models. The bottom row shows the segmentation masks generated from attention maps of slots. The numbers in the top right of each section are for the qualitative results of each model over each dataset (reported in $\%$).
  }
  \label{fig:qual}
\end{figure*}

\subsubsection{Kernels in ARK}
In our method, ARK is applied as a low-pass filter with a WNConv network to eliminate the noise and strengthen the object-like patterns in the attention maps.
In this study, we look into the alternatives for the WNConv.
Firstly, we compare our WNConv with the global smoothing scheme where we increase the temperature $\tau$ 
The logit values of an attention map are divided by $\tau$ so that the larger $\tau$ yield a more smoothed attention map.
Secondly, we apply a representative smoothing technique, Gaussian filter \cite{vernon1991machine}.
Lastly, we conduct experiments on a Convolutional (Conv) kernel without any constraints like weight normalization.

\cref{tab:ablation_ark} demonstrates the results of the possible kernels for ARK.
We observe that the SA models with high temperature and the Gaussian smoothing outperform the original SA.
This result implies that simple global and local smoothing can help a model boost its performance by erasing noises in the attention maps.
For the Conv kernel, we observe that the overall performance is worse than the other kernels.
We argue that the poor performance of the Conv kernel is incurred due to the high degree of freedom for the single-channel single-layer network.
In contrast, owing to the reduced degree of freedom, WNConv consistently performs well by recording high average values than the alternatives.

\begin{table}
\begin{center}
    \resizebox{0.95\linewidth}{!}{
    \begin{tabular}{lccc}
        \Xhline{3\arrayrulewidth}
        & \multicolumn{1}{c}{mIoU} & \multicolumn{1}{c}{ARI} &  \multicolumn{1}{c}{fg-ARI} \\ 
        \hline\hline & \multicolumn{3}c{CLEVR6} \\ \hline
        SA ($\tau=1$) \cite{locatello2020object} & 49.5 $\pm$ 20.5 & 63.0 $\pm$ 42.1 & 97.1 $\pm$ \phantom{0}1.6  \\
        SA ($\tau=2$) & 52.7 $\pm$ 24.6 & 66.4 $\pm$ 40.6 & 95.8 $\pm$ \phantom{0}2.8 \\
        Gaussian & 50.1 $\pm$ 14.9 & 75.6 $\pm$ 29.4 & 93.2 $\pm$ \phantom{0}2.4 \\
        Conv & 52.7 $\pm$ 19.6 & 76.8 $\pm$ 32.0 & 92.5 $\pm$ \phantom{0}2.2 \\
        WNConv & 64.1 $\pm$ \phantom{0}3.1 & 89.9 $\pm$ \phantom{0}2.2 & 95.3 $\pm$ \phantom{0}1.6 \\
        \Xhline{2\arrayrulewidth} & \multicolumn{3}{c}{CLEVRTEX} \\ \hline
        SA ($\tau=1$) & 22.2 $\pm$ \phantom{0}4.3 & 38.1 $\pm$ 12.5 & 52.1 $\pm$ \phantom{0}5.9 \\
        SA ($\tau=2$) & 25.6 $\pm$ \phantom{0}2.0 & 39.6 $\pm$ \phantom{0}3.7 & 54.9 $\pm$ \phantom{0}2.7 \\
        Gaussian & 26.0 $\pm$ \phantom{0}8.5 & 43.5 $\pm$ 14.6 & 55.9 $\pm$ 11.0 \\
        Conv & 24.8 $\pm$ \phantom{0}6.0 & 42.5 $\pm$ \phantom{0}9.7 & 54.3 $\pm$ 11.1 \\
        WNConv & 31.4 $\pm$ \phantom{0}6.6 & 55.6 $\pm$ 13.2 & 57.8 $\pm$ \phantom{0}7.7 \\
        \Xhline{2\arrayrulewidth} & \multicolumn{3}{c}{PTR} \\ \hline
        SA ($\tau=1$) & 17.6 $\pm$ 14.7 & 19.6 $\pm$ 29.8 & 44.5 $\pm$ 18.8 \\
        SA ($\tau=2$) & 34.3 $\pm$ 12.0 & 56.6 $\pm$ 26.5 & 50.0 $\pm$ \phantom{0}7.9 \\
        Gaussian & 20.6 $\pm$ 15.1 & 20.0 $\pm$ 29.9 & 53.8 $\pm$ 10.2 \\
        Conv & 12.4 $\pm$ \phantom{0}9.7 & 11.6 $\pm$ 13.1 & 32.1 $\pm$ 26.0 \\
        WNConv & 43.8 $\pm$ \phantom{0}3.0 & 62.3 $\pm$ 19.4 & 60.4 $\pm$ \phantom{0}3.2 \\ 
        \Xhline{2\arrayrulewidth} & \multicolumn{3}{c}{MOVi} \\ \hline
        SA ($\tau=1$) & 23.0 $\pm$ \phantom{0}9.8 & 25.9 $\pm$ 20.3 & 48.7 $\pm$ \phantom{0}7.0 \\
        SA ($\tau=2$) & 27.1 $\pm$ \phantom{0}5.5 & 28.7 $\pm$ 12.5 & 54.6 $\pm$ \phantom{0}1.8 \\
        Gaussian & 25.5 $\pm$ 10.8 & 33.7 $\pm$ 21.5 & 48.5 $\pm$ \phantom{0}7.1 \\
        Conv & 25.5 $\pm$ \phantom{0}8.8 & 28.2 $\pm$ 18.0 & 53.0 $\pm$ \phantom{0}2.4 \\
        WNConv  & 27.2 $\pm$ \phantom{0}6.1 & 33.2 $\pm$ 13.7 & 57.0 $\pm$ \phantom{0}3.7 \\ 
        \Xhline{3\arrayrulewidth}
    \end{tabular}
    }
    \end{center}
    \vspace{-1em}
    \caption{Results of ablation studies on the alternatives of the kernel for ARK (mean $\pm$ std for 10 trials, reported in $\%$). 
    SA stands for the baseline Slot Attention \cite{locatello2020object}. $\tau$ is a temperature coefficient in the attention mechanism.
    }
    \label{tab:ablation_ark}
\end{table}

\subsection{Analysis of Bleeding Issues}
In this section, we investigate the cases where the baseline, Slot Attention \cite{locatello2020object}, fails to prevent the bleeding issue and SLASH succeeds in that.
\cref{fig:qual} shows the results by the baseline and SLASH.
Here, we present our analysis of the failure cases for each dataset.

As depicted in \cref{fig:qual} (a), for CLEVR6, SA encounters the bleeding issue due to the simplicity of the background.
As shown in the top-left image, CLEVR6 only contains simple white backgrounds without any complicated pattern.
Since the background has almost no information, a model is likely to get into the trivial solution that every slot binds to the piece of the background.

As shown in \cref{fig:qual} (b), stripping frequently occurs for PTR.
The striping issue is a phenomenon where each slot is trapped into a simple and meaningless stripe pattern of an image.
We assume the striping issue occurs as a model tends to focus on positional embedding rather than object-related patterns that are difficult for the model to figure out.

CLEVRTEX contains a variety of complex backgrounds as shown in \cref{fig:qual} (c).
In SA, slots tend to be attracted by the explicit eye-catching patterns on backgrounds.
We argue that this phenomenon is attributed to the design of SA which focuses on the versatility towards domain- and task-agnostic models.
This design principle results in the lack of inductive biases and locational information for discovering objects rather than backgrounds.

As SLASH is designed to not only eliminate background noise and solidify object-like patterns in the attention map, but also encode the positional information into the slots, we observe that SLASH is robust against the aforementioned failure cases in various datasets.

\section{Conclusion and Limitation}
In this paper, we observed that OCL for single-view images has a stability issue that some training trials end up with having the bleeding issue.
We attributed this problem to the lack of inductive bias about the appearance of objects and additional cues like positional information.  
We presented a new OCL framework for single-view images, called SLASH, acting as a shepherd, guiding the slots to the correct destination without being distracted by the background noise. 
To accomplish this, we proposed two simple modules: ARK for smoothing the noise in the attention and IPPE for inducing positional information through a weak semi-supervision. 
Experimental results show the effectiveness of our method, which achieved strong and consistent results for stable and robust OCL.

Although our model shows impressive results on various challenging synthetic datasets, extending our method to real-world image datasets remains a problem and promising path for future work. 
There are several potential challenges to achieving real-world OCL: understanding backgrounds, controlling a large number of objects, handling the representation of an object having intricate shape and texture, and designing an efficient model that can process high-resolution images.
We expound on additional limitations of our study in the supplementary material.

\section{Acknowledgement}
This research was supported and funded by Artificial Intelligence Graduate School Program under Grant 2020-0-01361, Artificial Intelligence Innovation Hub under Grant 2021-0-02068, and the Korean National Police Agency. [Project Name: XR Counter-Terrorism Education and Training Test Bed Establishment/Project Number: PR08-04-000-21].


\newpage
{\small
\bibliographystyle{ieee_fullname}
\bibliography{egbib}
}

\end{document}